\documentclass[10pt,twocolumn,letterpaper]{article}

\usepackage{cvpr}      

\usepackage{graphicx}
\usepackage{amsmath}
\usepackage{amssymb}
\usepackage{booktabs}

\usepackage[pagebackref,breaklinks,colorlinks]{hyperref}

\usepackage[capitalize]{cleveref}
\crefname{section}{Sec.}{Secs.}
\Crefname{section}{Section}{Sections}
\Crefname{table}{Table}{Tables}
\crefname{table}{Tab.}{Tabs.}

\def\cvprPaperID{*****} 
\def\confName{CVPRW}
\def\confYear{2022}

\begin{document}

\title{A Tour of  Visualization Techniques for Computer Vision Datasets}

\author{Bilal Alsallakh\\
Meta
\and
Pamela Bhattacharya\\
Meta\\
\and
Vanessa Feng\\
Meta\\
\and
Narine Kokhlikyan\\
Meta\\
\and
Orion Reblitz-Richardson\\
Meta\\
\and
Rahul Rajan\\
Meta\\
\and
David Yan\\
Meta\\
}
\maketitle

\begin{abstract}
   We survey a number of data visualization techniques for analyzing Computer Vision (CV) datasets.
   These techniques help us understand properties and latent patterns in such data, by applying dataset-level analysis.
   We present various examples of how such analysis helps predict the potential impact of the dataset properties on CV models and informs appropriate mitigation of their shortcomings.
   Finally, we explore avenues for further visualization techniques of different modalities of CV datasets as well as ones that are tailored to support specific CV tasks and analysis needs.
\end{abstract}

\section{Introduction}
\label{sec:intro}

The majority of work on understanding CV algorithms and explaining their behavior and results focuses on analyzing the internals of CV models and revealing the features they learn during training.
On the other hand, less attention has been paid to analyzing CV datasets and their properties, despite the impact of these properties on the model behavior and on the learned features.
Existing work tends to rely on quantitative analysis~\cite{zendel2017analyzing} of these datasets, with limited focus on exploratory visual analysis.

Torralba and Efros~\cite{Torralba2011Unbiased} assessed the quality of various CV datasets.
Their comparative analysis illustrates different types of bias in these datasets which limit their generalizability and representativeness.
Likewise, various studies examined label imbalance issues in various CV datasets and their impact on different tasks, focusing on solutions to mitigate the bias associated with them. 
Fabbrizzi et al.~\cite{fabbrizzi2021survey} provide an excellent overview of these studies.
For example, Oksuz et al.~\cite{oksuz2020imbalance} studied different types of imbalance in object detection stemming from the interaction between foreground and background objects, their sizes, and locations.
Nevertheless, as noted by Wang et al.~\cite{revisetool_eccv} there is a lack of reusable tools and techniques to systematically surface different types of bias in CV datasets.
We also note that most techniques rely on quantitative analysis, and rarely leverage data visualization as a powerful means to analyze CV datasets.

We present visualization techniques that help analyze fundamental properties of CV datasets.
The goal of this analysis is to improve our understanding of these properties and how they can potentially impact CV models and algorithms.
Our contributions include:
\begin{itemize}
    \item Providing an overview of generally applicable visualization techniques for CV datasets.
    \item Demonstrating how the insights these techniques provide help expose potential shortcomings of CV datasets and explain CV model behavior.
\end{itemize}
In Section~\ref{sec:discussion} we explore further analysis opportunities that are not well supported by existing techniques.

\begin{figure*}[th!]
    \centering
    \includegraphics[width=\textwidth]{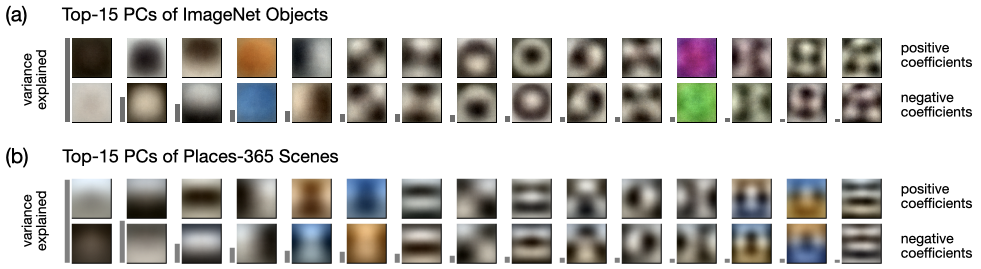}
    \caption{The top-15 principal components in two CV datasets ImageNet and Places365. In both cases, the images were resized to $40 \times 40$
    (a) The analysis is applied to the bounding boxes of ImageNet objects.
    (b) The analysis is applied to full scenes.}
    \label{fig:PCA_wholeimage}
\end{figure*}

\section{A Tour of Visualization Techniques}

This section surveys a number of visualization techniques that can be applied to CV datasets.
For each technique, we demonstrate example insights in popular datasets, as well as possible implications of these insights.

\subsection{Pixel-Level Component Analysis}

Principal Component Analysis (PCA) is widely used as a nonparametric method for dimensionality reduction.
When used for analysis and visualization purposes, oftentimes, the data points are projected along the top-2 or top-3 principal components (PCs) and visualized using 2D or 3D plots.
Such plots are rather limited when PCA is applied in the flattened pixel space, which is usually very high-dimensional.
Furthermore, a relatively small number of data points can be intelligibly visualized in 2D plots, especially when image icons are displayed. 

Fortunately, it is possible to visualize the PCs of a flattened pixel space directly as visual entities, without projecting data points on them.
This is because each PC is an eigenvector defined as a linear combination of the dimensions of the mean-centered pixel space.
Each value in this vector corresponds to one color channel of a unique pixel and can be either positive or negative.
A pictorial representation of these values can hence visualize the image features to which the respective PC is sensitive.

We propose visualizing each PC by means of two images, one consisting of the positive values of its eigenvector (with the negative ones treated as zero), and the other consisting of the inverted negative values (with the positive ones treated as zero).
This decomposes the visual representation into two simpler parts, each showing which pixels tend to have high value at one end of the PC direction in the data space.
This reveals subtle details in both images that are not always visible when visualizing the eigenvector values in a single image.
For example, this reveals that the variance explained by the 4-th PC in Figure~\ref{fig:PCA_wholeimage}a is between orange-dominated vs. blue-dominated images.
To the side of the two images, we depict a gray bar whose length represents the eigenvalue of each PC.
Accordingly, the visualization of the pixel space hence consists of the two extremes of each PC, in addition to a bar chart of the eigenvalues.

We next present two different modes to apply PCA to image data based on the unit of analysis, and elaborate on the insights they provide in the dataset.
The first mode operates on whole images while the second operates on small patches in these images.

\subsubsection{PCA on Whole Images}

To apply PCA to a dataset whose items are images, these images must be of the same size and color depth.
This allows treating the dataset as a high-dimensional table.
Furthermore, PCA requires the number of data points $n$ to be larger than the number of dimensions $p$ in the table to allow consistent estimation of the subspace of maximal variance.
Depending on the dataset and the goal of analysis, recommendations for the ratio between sample size and number of dimensions vary between 1 and 100, with higher ratios usually recommended for very high-dimensional datasets.

The above requirements are generally straightforward to meet for the purpose of analyzing and understanding vision datasets by applying suitable crop and resize operations.
While downscaling the images can compromise fine-grained details, applying PCA to downscaled images is useful to analyze global factors of variations among them.
Furthermore, with small images the computation overhead is significantly smaller than with large ones~\footnote{The  time complexity of standard PCA is $O(p^3 + p^2 \cdot n)$. Faster algorithms exist for iteratively approximating the top PCs.}.
Accordingly, we recommend starting the analysis with relatively small images, and validating the axes found by repeating the analysis on larger resolutions while avoiding overfitting.

In Figure~\ref{fig:PCA_wholeimage}a, the PCs were computing using the ImageNet validation set, which contains $n = 50.000$ data points.
We first crop each image to the bounding box of the target object and then resize the image further to $40 \times 40$ pixels, resulting in $k = 4800$ dimensions and hence a ratio of $n / k = 10.4$.
As evident in Figure~\ref{fig:PCA_wholeimage}a, the top-5 PCs correspond to low-frequency features that together explain about $40\%$ of the variance in the dataset. 
\begin{itemize}
  \setlength\itemsep{0em}
    \item 
    The 1st PC differentiates between dark and bright object images,
    explaining about $24\%$ of the variance.
    \item 
    The 2nd PC differentiates between dark objects on a bright background, and the opposite case.
    \item 
    The 3rd PC is sensitive to whether the brightness is concentrated in the upper or bottom parts of the image.
    \item 
    The 4th PC differentiates based on orange and sky-blue tints, two opposing colors in the RGB cube.
    \item
    The 5th PC is sensitive to whether the brightness is concentrated in the left or right parts of the image.
\end{itemize}
The remaining PCs correspond to increasingly higher-frequency features in the spatial dimensions.
One exception is the 12th PC, which differentiates based violet and green tints, which are also opposing colors in the RGB cube.
Interestingly, this PC explains only $1.24\%$ of the variance, while the orange-blue PC explains $3.3\%$ of it.
This suggests that perturbations along these dimensions might have varying impact on the learned models, as we demonstrate in Section~\ref{sec:how_useful_is_ca}.
In Figure~\ref{fig:PCA_wholeimage}b, the same analysis is applied to the Places365 validation set.
Notice how:
\begin{itemize}
\item
Horizontal bands dominate the top PCs, such as the 3rd and 7th PCs, while vertical bands are less salient and are often mixed with the blue-orange dimension.
This reflects the nature of scene-centric datasets, in contrast with the object-centric ImageNet dataset.
\item
The green-violet axis is not among the top-15 components of Place-365, while the blue-orange dimension is correlated with four of them.
This is because scene images are rarely dominated by violet color, which limits the variance explained by the green-violet axis.
\end{itemize}



\subsubsection{PCA on Image Patches}

Instead of treating images as the unit of analysis, it is possible to apply PCA to image patches.
This mode of analysis has been used in the literature to compare images~\cite{shan2008looking}, denoise them~\cite{deledalle2011image}, and to extract features for recognition tasks~\cite{jiang2017patch}.
Furthermore, it has been used to understand the statistics of natural images and to link them with certain mechanisms of human vision~\cite{hancock1992principal, ruderman1998statistics, eichhorn2009natural, Hyvrinen2009book}.

Applying PCA to image patches eliminates the need for resizing the images.
When used to understand the influence of the data on the model learned, the patch size can be selected to match how the model processes the images.
For example, when convolutional neural networks are used, the size can be selected to match the filter size of the first convolutional layer.

\begin{figure}[h!]
    \centering
    \includegraphics[width=\linewidth]{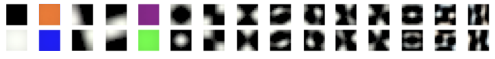}
    \caption{The top-15 principal components in 3.1 million image patches of size $11 \times 11 \times 3$, sampled from ImageNet.}
    \label{fig:PCA_ImageNet_Patch}
\end{figure}
\begin{figure}[h!]
    \centering
    \includegraphics[width=0.8\linewidth]{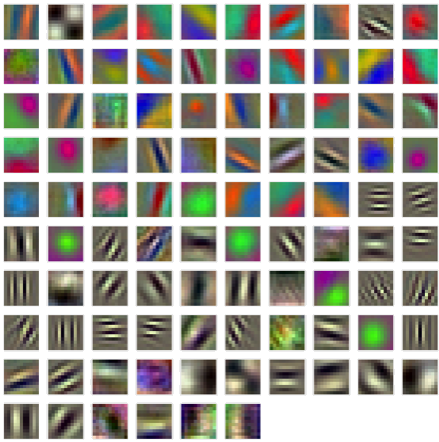}
    \caption{The filters of the first convolutional layer in AlexNet. Each filter is of size $11 \times 11 \times 3$.
    The color intensity is normalized individually per filter, to emphasize the color components.}
    \label{fig:AlexNet_ImageNet_Filters}
\end{figure}

Figure~\ref{fig:PCA_ImageNet_Patch} shows the PCA results computed for 3.1 million  patches of size $11 \times 11$, randomly sampled from the ImageNet validation dataset at random locations spanning the entire image area.
The size corresponds to the filter size of the first convolutional layer in AlexNet~\cite{krizhevsky2012imagenet}. 
Figure~\ref{fig:AlexNet_ImageNet_Filters} depicts the $96$ filters learned by this model when trained to classify ImageNet.
Notice how the color filters learned by AlexNet largely follow the two color dimensions found among the top PCs.

\subsubsection{Independent Component Analysis (ICA)}

PCA is designed to maximize the variance explained by the top PCs it computes.
When applied to images, these PCs hence tend to involve all pixels in order to find directions that maximize this variance.
Accordingly, the visual features that correspond to these directions are rather global, as evident in Figure~\ref{fig:PCA_wholeimage}.
Moreover, these directions can be correlated.

\begin{figure}[h!]
    \centering
    \includegraphics[width=\linewidth]{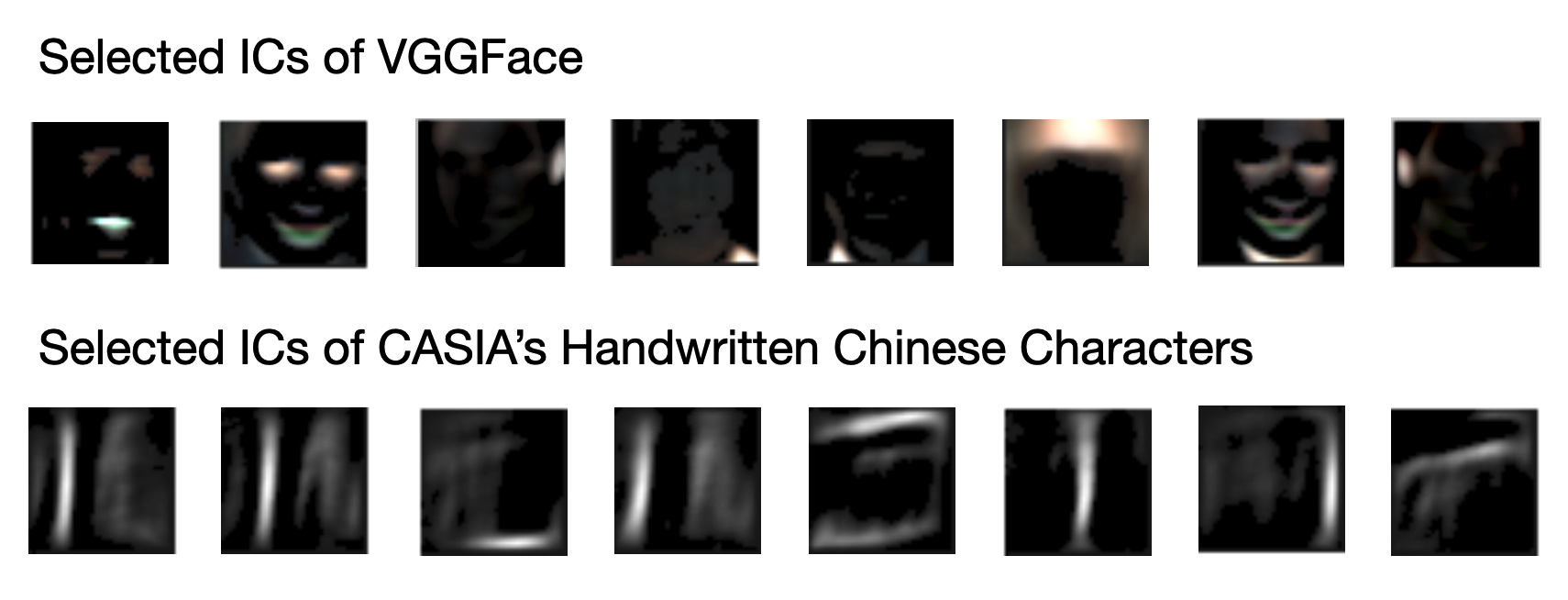}
    \caption{Selected independent components in two datasets, VGG Face, and CASIA's Handwritten Chinese Characters.
    Notice how ICA can identify localized features.}
    \label{fig:ICA_Examples}
\end{figure}

ICA~\cite{HYVARINEN2000411} offers an alternative component analysis that aims to maximize the \textit{independence} between these components, instead of the variance they explain.
When applied to images, the independent components (ICs) found can hence focus on localized features that are uncorrelated.
Figure ~\ref{fig:ICA_Examples} shows examples of ICs computed for two datasets, VGG Face~\cite{Parkhi15} and the CASIA Chinese Offline Handwritten Characters~\cite{liu2011casia}.
The ICs were computed using Fast ICA \cite{hyvarinen1999fast}, applied to images of size $40 \times 40$.
Notice how the IC components are localized and highly disentangled from each other.

When applied to image patches sampled from natural image collections, ICA was shown to learn features that resemble Gabor functions~\cite{Hyvrinen2009book}.
The gray-scale filters in Figure~\ref{fig:AlexNet_ImageNet_Filters} also resemble Gabor functions.
This suggests that these filters aim to maximize the independence between their features.
It is useful to reduce the dimensionality of the data space using PCA, before applying ICA, in order to discourage ICA from overfitting noise patterns.

\subsubsection*{How useful is component analysis?}
\label{sec:how_useful_is_ca}
Exploring the top PCs in the pixel space is helpful to understand which image features are behind significant variations in the dataset, and accordingly predict their potential importance for the model.
For example, it is evident in Figure~\ref{fig:PCA_wholeimage}a and in Figure~\ref{fig:PCA_ImageNet_Patch} that the blue-orange direction explains more variance than the green-violet direction in the ImageNet pixel space.
Accordingly, we suspect that ablating the blue color channel will likely have more impact on ImageNet classifiers than ablating the green channel.
To verify this assumption, we ablate each of the three color channels in the input of ResNet-18 and report its top-1 accuracy on ImageNet validation set in Table~\ref{table:imagenet}.
We perform ablation either by replacing the color channel with the mean of the other two channels, or with  the mean of all three channels (i.e. a basic gray image).

\begin{table}[!h]
\small
\caption{Top-1 accuracy of ResNet-18 trained on ImageNet when different color channels are masked in the validation set. The baseline accuracy is $69.68\%$}
\label{table:imagenet}
\centering
\bgroup
\def\arraystretch{1.5}
\setlength{\tabcolsep}{8pt}
\begin{tabular}{ccc}
     Mask channel w.: & Mean of other channels & Gray image \\
    \hline
    Red & $56.50\%$  & $63.70\%$  \\ 
    Green & $62.12\%$  & $66.57\%$  \\ 
    Blue & $60.56\%$  & $64.93\%$  \\ 
\end{tabular}
\egroup 
\end{table}
As evident in Table~\ref{table:imagenet}, ablating the blue channel indeed has a higher impact on performance, compared with the green channel.
Moreover, ablating the red channel has the highest impact, since it is strongly involved both in the violet and in the orange PC extremes.
These results generalize to a variety of models as well as to CIFAR-10.
This illustrates the value of understanding the major dimensions of variation in the pixel space.

\subsection{Spatial Analysis}

CV datasets usually contain annotations of various objects in their images, typically in the form of bounding boxes or segmentation masks.
Visualizing these boxes and masks for an entire dataset is helpful to understand how the corresponding objects are spatially distributed.
\begin{figure}[h!]
    \centering
    \includegraphics[width=\linewidth]{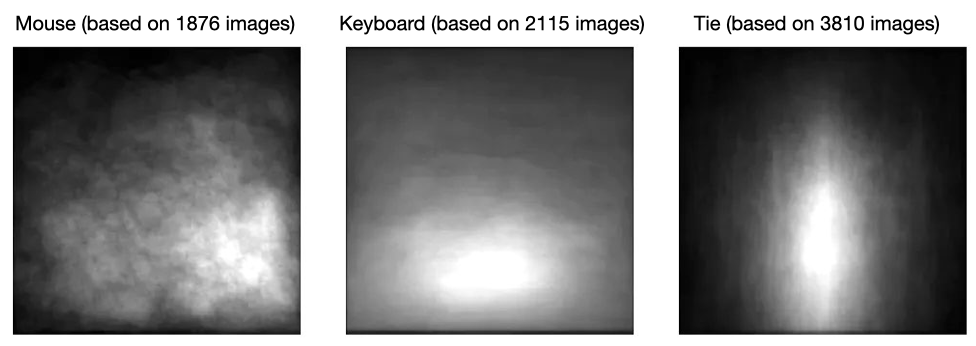}
    \caption{The spatial distribution of three object categories in the MS COCO dataset, obtained by aggregating their masks.}
    \label{fig:COCO_spatial}
\end{figure}
Figure~\ref{fig:COCO_spatial} depicts the spatial distribution of three object categories in the MS COCO dataset~\cite{lin2014microsoft}.
Each plot aggregates the mask images of dataset samples that contain the corresponding object.
The aggregation is performed by resizing the mask images to $640 \times 640$ pixels, summing up the object pixels in these mask into an aggregate image, and normalizing the resulting image to the range $[0, 255]$.
It is also possible to analyze further spatial relationships in CV datasets, such as co-occurrence and  adjacency relationships between various categories in semantic segmentation~\cite{revisetool_eccv}.

\subsubsection*{How useful is this analysis?}

\begin{figure}[!h]
    \centering
    \includegraphics[width=\linewidth]{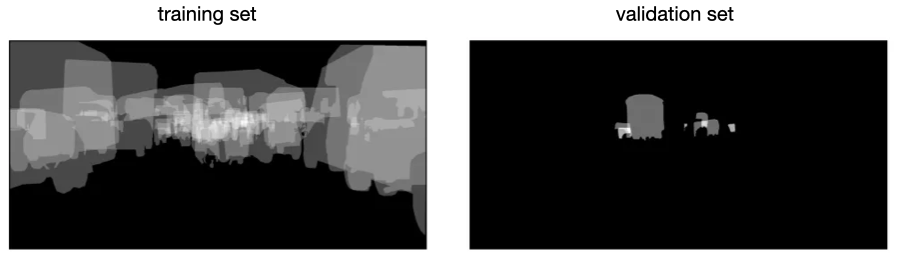}
    \caption{The spatial distribution of the \texttt{caravan} category in two splits of the CityScapes dataset.}
    \label{fig:CityScapes_Cravan}
\end{figure}

Analyzing the spatial distribution of object categories helps uncover potential shortcomings of a CV dataset.
Figure~\ref{fig:CityScapes_Cravan} illustrates how the spatial distribution of \texttt{caravan} category in CityScapes~\cite{cordts2016cityscapes} varies significantly between the training and validation sets.
Likewise, Gauen et al~\cite{gauen2017comparison} show differences in how person images are spatially distributed in seven popular datasets, which helps evaluate transferability between them.
The analysis can also help assessing whether popular data augmentation methods are suited to mitigate any skewness in the spatial distribution.
For example, the commonly used horizontal flipping helps create training sample that contain a computer mouse both on the left side and on the right side of the image.
However, as evident in Figure~\ref{fig:COCO_spatial}, the samples remain predominantly in the lower part of the image.
Moreover,  horizontal flipping might negatively impact images with orientation-sensitive content such as written text~\cite{Torralba2011Unbiased}.

\begin{figure}[th!]
    \centering
    \includegraphics[width=\linewidth]{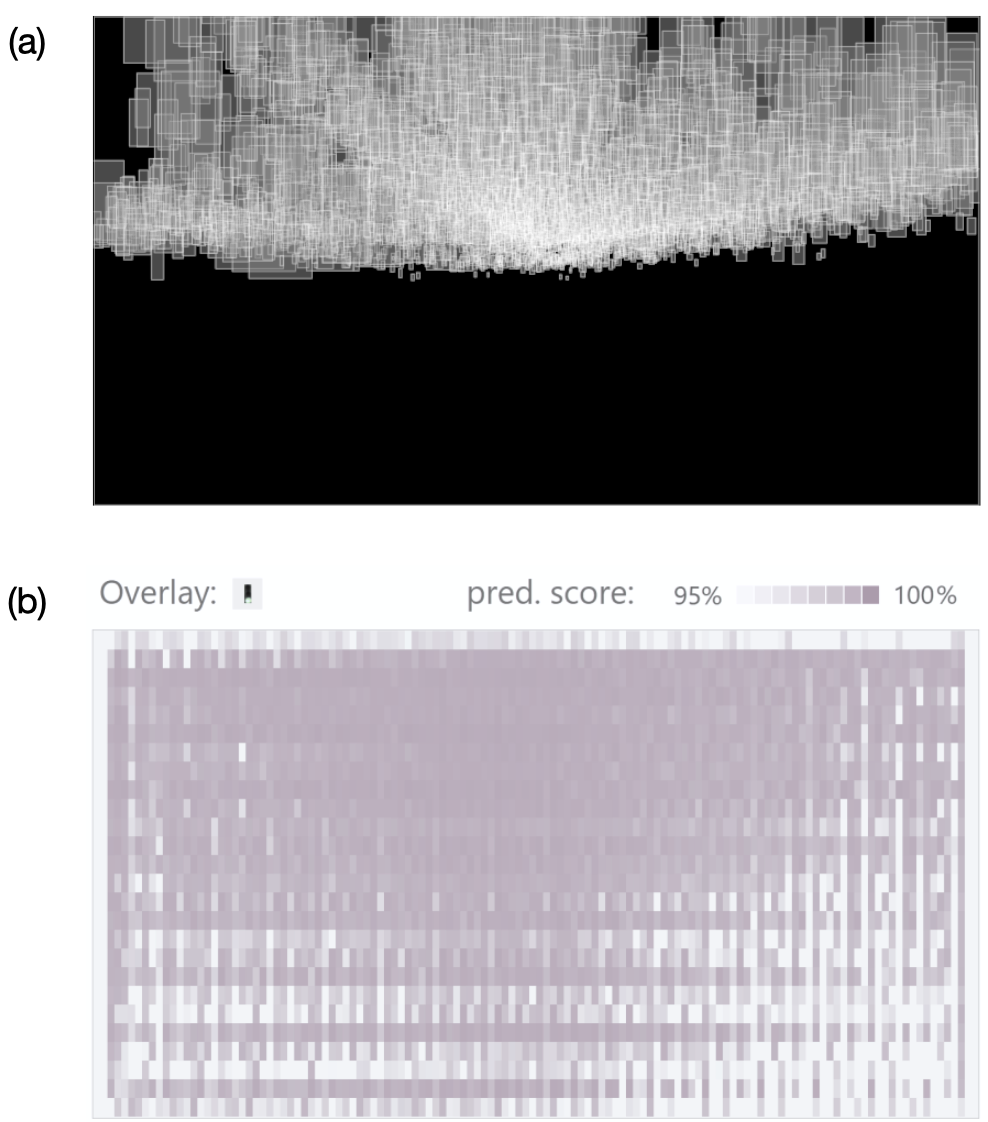}
    \caption{
    The impact of spatial distribution on RCNN-based small object detection.
    (a) The distribution of traffic light bounding boxes in the BSTLD dataset~\cite{behrendt2017deep}.
    (b) The prediction scores for a traffic light stimulus when placed at various locations in a blank image, computed via Faster RCNN. (Figure adapted from~\cite{alsallakh2021visual}).
    }
    \label{fig:TL_Distribs}
\end{figure}

To appreciate the impact of spatial distribution, we present an example of a traffic light detector that was shown to learn and overfit the spatial distribution of traffic lights in the training set~\cite{alsallakh2021visual}.
This distribution is depicted in Figure~\ref{fig:TL_Distribs}a.
Figure~\ref{fig:TL_Distribs}b depicts the detection score the the model computes a traffic light stimulus placed in a blank image.
The score is evidently higher if the stimulus is present in a position that has a high density of training samples.
Such overfitting is possible as convolutional networks are able to encode position information~\cite{islam2020much}, with padding being a major source of this information~\cite{kayhan2020translation}.
 
\subsection{Average Image Analysis}

\begin{figure}[!h]
    \centering
    \includegraphics[width=0.9\linewidth]{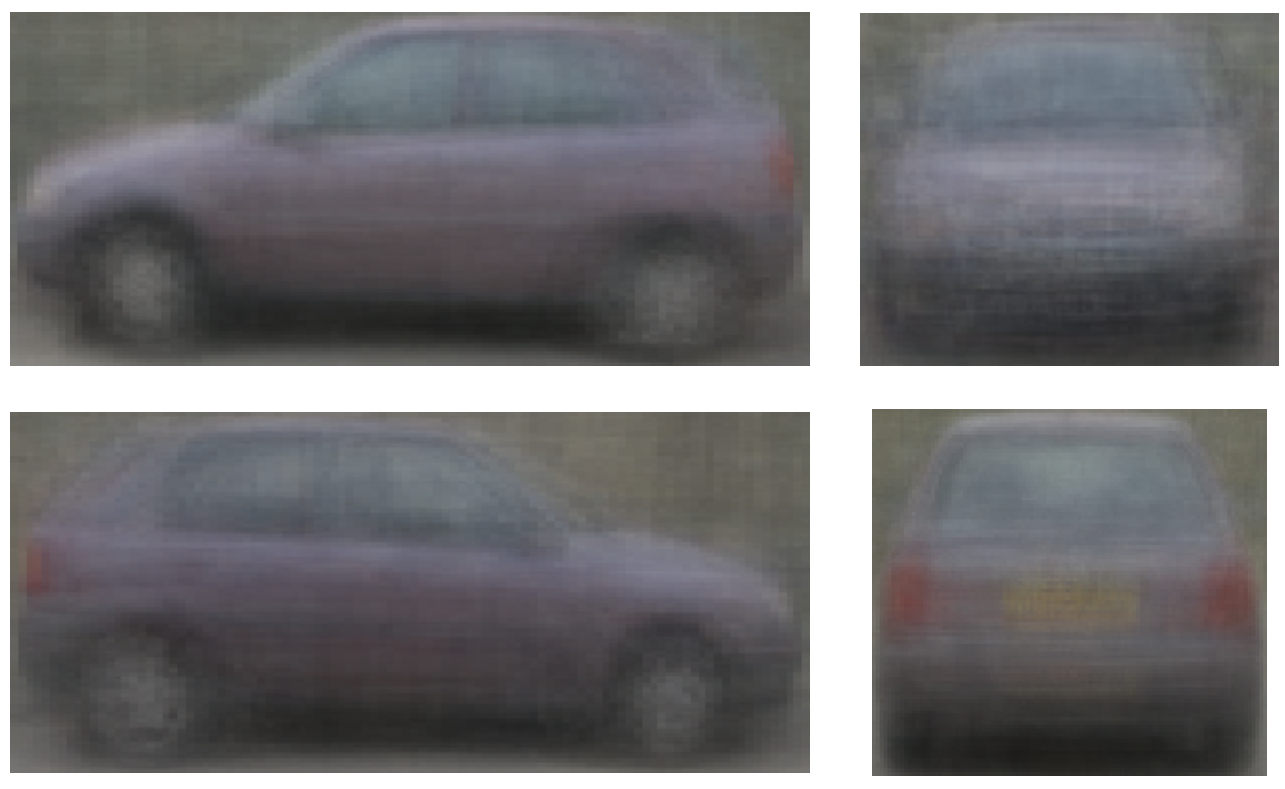}
    \caption{
    Average images of the car category in PASCAL VOC 2006~\cite{pascal-voc-2006}, computed for different poses (Figure adapted from \url{https://cs.cmu.edu/~tmalisie/pascal}).
    }
    \label{fig:VOC_avg_images}
\end{figure}

Averaging a collection of images offers a useful visual summary that helps detect various issues in CV datasets~\cite{Ponce2006}.
Average images can also be used to compare subsets of images of the same nature and semantics, e.g., to trace how portrait photos change over time~\cite{ginosar2015century} or to compare hand-drawings of the same object across different cultures~\cite{martino2017forma}.
Interactive clustering exploration is often helpful to select subsets that represent representative manifestations or interesting outliers~\cite{zhu2014averageexplorer, martino2017forma}.
Likewise, when computing average images of selected objects, it is helpful to group the individual images based on the object pose, as illustrated in Figure~\ref{fig:VOC_avg_images}.
\subsubsection*{How useful is this analysis?}
Computing average images per class~\cite{torralba2003statistics} can reveal visual cues in classification datasets that classifiers can use as ``shortcuts'' instead of learning robust semantic features.
Analyzing these cues helps in designing dataset augmentations that prevent models from using them.
In addition, the mean image of ImageNet was used to investigate the source of weight banding effects~\cite{petrov2021weight} in certain models that use global average pooling~\cite{alsallakh2021debugging}.
The authors demonstrate further applications of image averaging to analyze and debug the internals of convolutional networks.

\subsection{Metadata and Content-based Analysis}

\begin{figure}[h!]
    \centering
    \includegraphics[width=\linewidth]{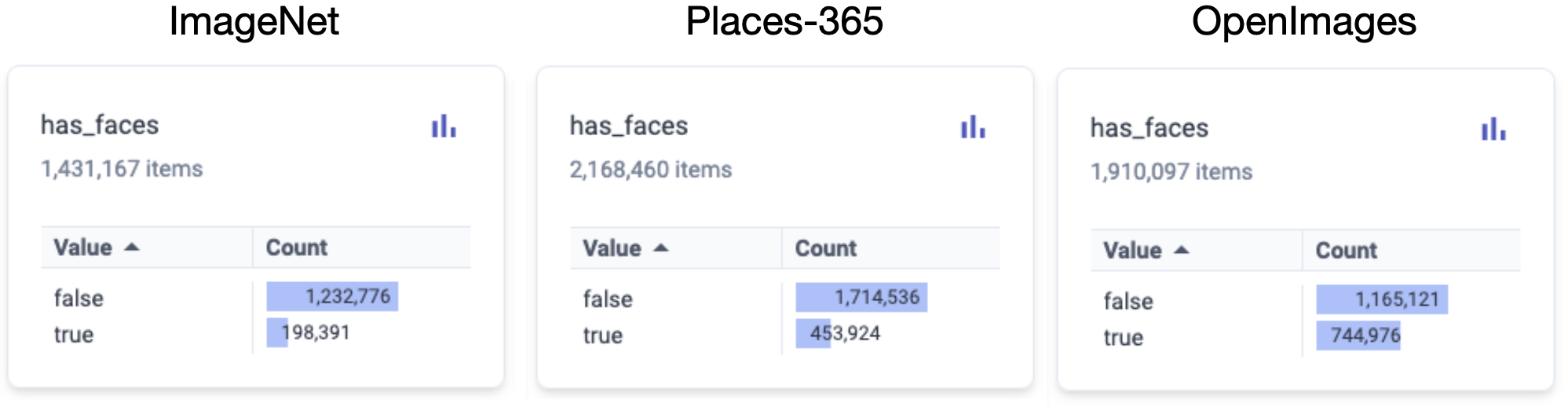}
    \caption{
    Examples of metadata summaries in Google's Know Your Data, extracted from three popular datasets.
    Notice how OpenImages V4~\cite{kuznetsova2020open} has the highest proportion of images that contain faces, while ImageNet has the lowest such proportion. 
    }
    \label{fig:Metadata_example}
\end{figure}

Google's Know Your Data\footnote{Know Your Data: \url{http://knowyourdata.withgoogle.com}} offers means to understand datasets with the goal of improving data quality and mitigating bias.
Users can analyze metadata about the images such as their aspect ratios and resolution, 
extracted signals such as image quality and sharpness, or various high-level features such as the presence of faces (Figure~\ref{fig:Metadata_example}).
Likewise, Google's Facets\footnote{Facets: \url{http://pair-code.github.io/facets/}} offer a zoomable interface to explore large datasets using thumbnails, with several possibilities to split and arrange the data points into scatter plots, based on their metadata.

It is often possible to partially analyze the geographic distribution of vision datasets based on available location information in their metadata.
Likewise, exploring datasets based on their temporal information is useful to assess their appropriateness for the target task.
Such analysis can account for several aspects of the time dimension such as the time of day, time of year, and overall recency of the samples.
Unavailable metadata can sometimes be recovered using specialized techniques, e.g. for dating images~\cite{Palermo2012, kim2013time} or for approximating the geolocation~\cite{doersch2012makes, weyand2016planet}.

\subsubsection*{How useful is this analysis?}
\begin{figure}[h!]
    \centering
    \includegraphics[width=0.85\linewidth]{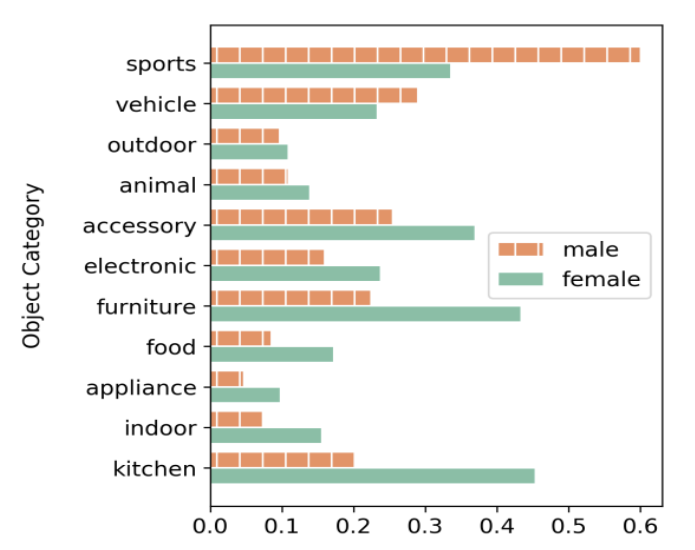}
    \caption{
    Content-based analysis of MS COCO using the REVISE tool~\cite{revisetool_eccv} focusing on intersectional gender bias in scenes that contain persons alongside different objects. 
    }
    \label{fig:Content_example}
\end{figure}

Geolocation analysis helps assess the diversity of a dataset and explore different manifestations of targeted classes and features, in order to guide the curation and labeling of representative datasets~\cite{de2019does, shankar2017no, revisetool_eccv}.
Likewise, exploring datasets based on their temporal information is useful to assess their appropriateness for the target task.
The REVISE tool~\cite{revisetool_eccv} demonstrates a variety of examples of how the above ideas can be applied to analyze various types of biases in vision datasets along three dimensions: objects, attributes, and geography. Figure~\ref{fig:Content_example} shows an example of such analysis.

\subsection{Analysis using Trained Models}

The analysis techniques presented so far can be applied directly to datasets, without requiring a model trained on them.
When available, such models offer a lens into the dataset that can reveal potential biases or deficiencies in it.

A variety of analysis methods have been developed to understand the features learned by CV models, such as feature saliency in a given input~\cite{bach2015pixel, selvaraju2017grad, sundararajan2016gradients, zeiler2014visualizing}, input optimization~\cite{yosinskiunderstanding, NIPS2016_5d79099f, olah2017feature}, and concept-based interpretation~\cite{ghorbani2019towards, kim2018interpretability}.
While these techniques primarily focus on analyzing  model behavior, they were shown to be useful for identifying dataset issues.
For example  dumbbell images in ImageNet often contain arms~\cite{mordvintsev2015inceptionism}, leading to pictures of arms being classified as \texttt{dumbbell}.
Likewise, specialized saliency techniques developed for video recognition models can show biases in video datasets \cite{feichtenhofer2020deep}.
Shared Interest~\cite{boggust2021shared} and Activation Atlas~\cite{carter2019activation} enable systematic analysis of these issues on the dataset level.


\begin{figure}[h!]
    \centering
    \includegraphics[width=0.8\linewidth]{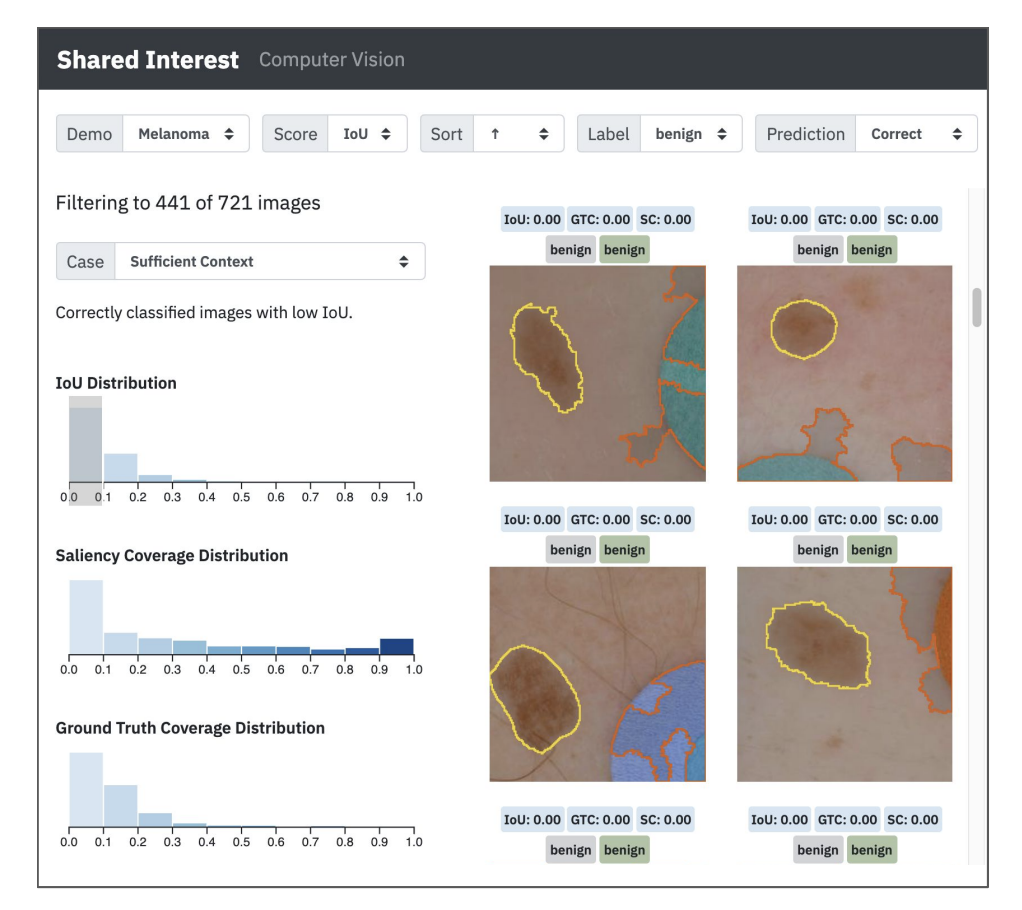}
    \caption{
    The Shared Interest UI~\cite{boggust2021shared} showing how well human annotations in the ISIC dataset~\cite{codella2019skin} match salient detection features.
    }
    \label{fig:shared_interest}
\end{figure}
Furthermore, various techniques use prediction results as a proxy to analyze dataset properties. 
For example, analyzing these results during training can reveal inherent sample difficulty~\cite{hacohen2020let, meding2022trivial}.
Likewise, analyzing error patterns in classification datasets using confusion matrices helps reveal latent hierarchical structures that govern their classes~\cite{deng2009imagenet, bilal2017convolutional}.

Finally, trained models can be used to compute embeddings that are suited to project the datasets into a 2D plot based on semantic features. Such a plot helps identify potential patterns and outliers in the dataset, especially when aided with interactive exploration.

\subsubsection*{How useful is this analysis?}
Difficulty analysis reveals the dichotomous nature of various CV datasets~\cite{meding2022trivial}.
For example, the majority of samples in ImageNet can either be correctly classified after a few training epochs, or continue to be misclassified throughout the training process.
Such insights are very helpful to understand the behavior of different learning paradigms and architectures, and how they are impacted by inherent issues in the training data.
\begin{figure*}[ht!]
    \centering
    \includegraphics[width=0.87\textwidth]{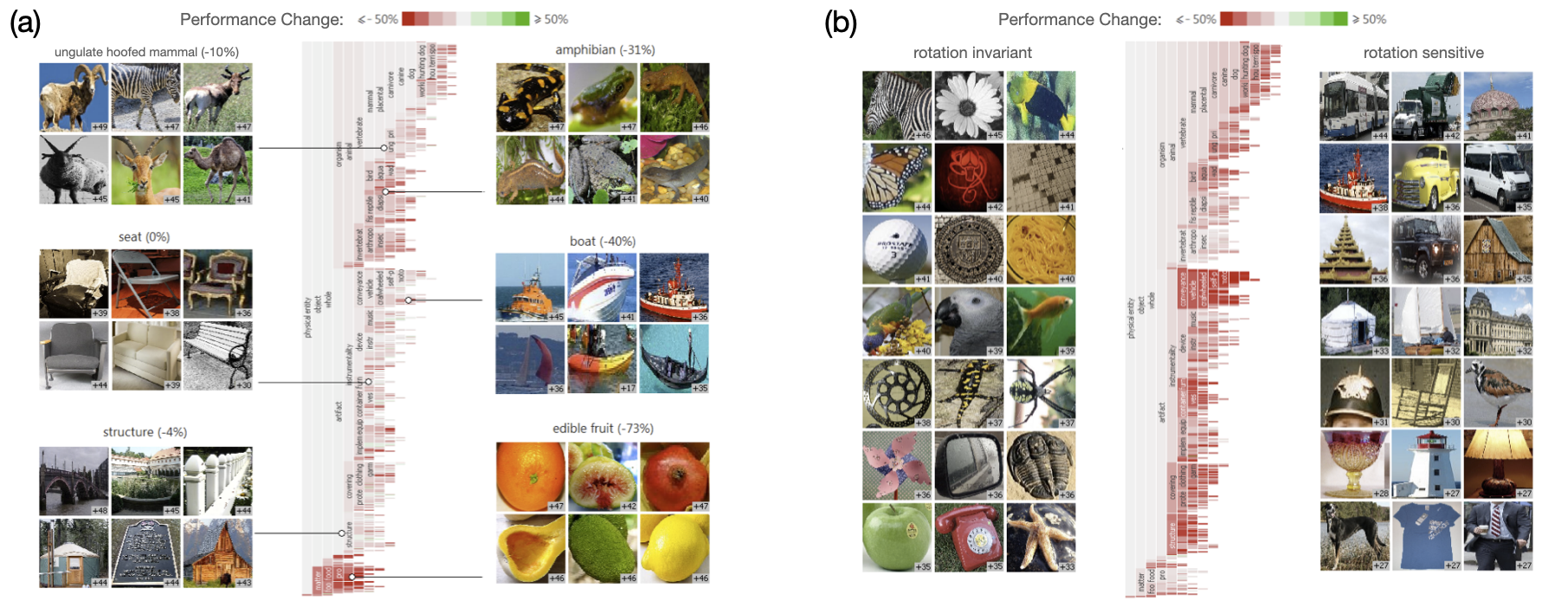}
    \caption{
    The effect of two perturbations on ImageNet classification accuracy, visualized for different classes and for their hierarchy~\cite{bilal2017convolutional}.
    The vertical dimension represents the classes, and the vertical icicle plot visualizes their hierarchy.
    Color encodes which classes and groups in the hierarchy are impacted most when their images are (a) converted to grayscale (b) rotated by $90^{\circ}$ degrees.
    }
    \label{fig:hierarchy}
\end{figure*}
Likewise, comparing human annotations with feature attribution results (Figure~\ref{fig:shared_interest}) was shown useful in revealing ambiguities and shortcoming of these annotations~\cite{boggust2021shared}.
Finally, visualizing latent hierarchical structures in classification datasets helps reveal the properties of different groups of classes, such as their reliance on color (Figure~\ref{fig:hierarchy}a) or orientation (Figure~\ref{fig:hierarchy}b).

\section{Discussion}
\label{sec:discussion}

Visual analysis offers distinctive opportunities to improve our understanding of CV datasets.
It can provide insights into their properties that are not always obvious or well understood.
The techniques presented focus almost exclusively on image datasets.
Here we discuss opportunities for further research on visualizing CV datasets.

\paragraph{Video, Point Cloud, 3D, and Keypoint Datasets}
Generally, there is a lack of dataset-level visualizations that are applicable to modalities in CV other than images.
Nevertheless, many of the techniques we presented can be extended to support these modalities.
Hyvrinen demonstrated how ICA helps understand spatiotemporal features in image sequences~\cite{Hyvrinen2009book}.
Likewise, t-SNE plots~\cite{van2008visualizing} are often used to visualize video embeddings in order to analyze the separability and overlaps between different classes~\cite{poorgholi2021t, tran2015learning}.
Dedicated visualization techniques can be further designed to reveal structures and patterns in non-image CV datasets based on the rich nature of their data modalities.

\paragraph{Task-driven Visual Analysis}
Besides supporting specific data modalities, the design of new visualization techniques should also take into account the target CV task.
For example, image classification datasets are different in nature that image segmentation datasets.
The corresponding visualizations can hence assign visual primacy in their designs to different facets, such as the class hierarchy in Figure~\ref{fig:hierarchy} and the spatial distribution in Figure~\ref{fig:COCO_spatial}.

\paragraph{Visually Linking Model Results with Dataset Properties}
Such an analysis can be useful to explain model behavior and find a suitable mitigation for observed shortcomings.
Figure~\ref{fig:TL_Distribs} demonstrates how the spatial distribution of the training set dictates object predictability at different locations.
Likewise, Figure~\ref{fig:hierarchy} demonstrates how different groups of classes vary in their robustness to different perturbations.

\paragraph{Comparing Datasets and Subsets Thereof}
Such comparisons have various applications, e.g., to assessing domain suitability in transfer learning and analyzing distribution shift~\cite{wiles2021fine}.
We presented two simple examples of comparing specific features between two image datasets (Figure~\ref{fig:PCA_wholeimage}) or between two splits of CityScapes (Figure~\ref{fig:CityScapes_Cravan}).
Further techniques can be designed to visualize similarities and differences between datasets e.g. by exposing the internals of similarity metrics.


\section{Conclusion}

Visualization is a powerful means for understanding CV datasets.
We presented a number of visualization techniques that enable analyzing various pieces of information in these datasets.
These techniques support dataset-level analysis to reveal latent patterns in the data and help us understand its properties.
For each technique, we demonstrated example insights it can enable in popular datasets.
We also provided application scenarios of how these insights help us understand the impact of dataset properties on CV models.
We finally explored a few avenues of how visualization can support further research on dataset understanding, towards robust and transparent CV.

{\small
\bibliographystyle{ieee_fullname}
\bibliography{main}
}

\end{document}